\documentclass{egpubl}
\usepackage{eg2026}
 
\ShortPresentation      
\usepackage[T1]{fontenc}
\usepackage{dfadobe}  
\usepackage{multirow}
\usepackage{booktabs}
\usepackage{amsmath}
\usepackage{amssymb}
\BibtexOrBiblatex

\electronicVersion
\PrintedOrElectronic
\ifpdf \usepackage[pdftex]{graphicx} \pdfcompresslevel=9
\else \usepackage[dvips]{graphicx} \fi

\usepackage{egweblnk} 

\title[RAW: Robust Avatar Watermarking]%
      {RAW: Robust Avatar Watermarking - Benchmarking and Baseline}

\author[J. Parry, J. Saunders \& V. Namboodiri]
{\parbox{\textwidth}{\centering Jack Parry$^{1}$\orcid{0009-0007-1474-0201},
Jack Saunders$^{1}$\orcid{0000-0001-8894-058X},
and Vinay Namboodiri$^{1}$\orcid{0000-0001-5262-9722}
\\$^{1}$University of Bath, UK}}

\begin{document}

\maketitle
\begin{abstract}
Digital avatar watermarking presents unique challenges: avatars are routinely post-processed with background replacement, reframing, and format conversion before deployment. We introduce \textbf{RAW} (Robust Avatar Watermarking), a benchmark comprising 50 synthetic avatar videos from 5 commercial providers and 6 attacks simulating real-world avatar workflows. Evaluating 7 existing methods reveals that avatar-specific attacks such as background removal significantly degrade watermark recovery. We propose \textbf{WALT} (Watermarking Avatars with Learned Textures), which embeds watermarks in UV texture space via 3D face reconstruction. WALT achieves the highest robustness to zoom attacks (92.4\%) while maintaining strong performance on background removal (95.6\%). We release our benchmark to facilitate research into avatar-specific watermarking.



\end{abstract}

\section{Introduction}
\label{sec:intro}

Digital avatars now form a multi-billion dollar industry~\cite{synthesia, heygen, tavus, d-id, vidnoz}. However, serious ethical and legal concerns accompany this success. Watermarking is essential to mitigating these, as recent legislation recognises. For example, the EU AI Act~\cite{eu-ai-act} states:

\begin{quote}
\textit{Providers [...] generating synthetic audio, image, video or text content, shall ensure that the outputs of the AI system are marked in a machine-readable format and detectable as artificially generated or manipulated. Providers shall ensure their technical solutions are effective, interoperable, robust and reliable.}
\end{quote}

A naive solution would be to attach metadata to generated videos, but metadata is easily stripped. A more robust approach is watermarking, which encodes binary data directly into the media. By embedding information such as the generating user or method, provenance can be recovered in cases of misuse.

However, existing watermarking methods are not designed for avatar workflows. Generated avatars are routinely post-processed before deployment: backgrounds are replaced for compositing into new scenes, videos are cropped or zoomed to reframe the subject, and formats are converted with varying compression levels. These transformations can destroy watermarks embedded by general-purpose methods.

To address this gap, we make three contributions. First, we introduce the \textbf{RAW Benchmark} (Robust Avatar Watermarking), comprising a novel dataset of 50 synthetic avatar videos from 5 commercial providers and a suite of 6 attacks designed to simulate real-world avatar post-processing. Second, we evaluate 7 existing watermarking methods on this benchmark, revealing that avatar-specific attacks pose significant challenges for current approaches. Third, we propose \textbf{WALT} (Watermarking Avatars with Learned Textures), a baseline that embeds watermarks in UV texture space using 3D face reconstruction. By operating in facial geometry space rather than pixel space, WALT achieves strong robustness to background removal and zoom while maintaining competitive performance on standard attacks.

A key finding is that simply restricting watermarks to the face region (which we try as a simple extension to VideoSeal \cite{videoseal}) improves robustness to avatar-specific attacks but catastrophically fails under video compression. WALT's learned texture approach recovers compression robustness while preserving the benefits of face localisation.

\begin{figure*}[th]
    \centering
    \setlength{\tabcolsep}{4pt}
    \begin{tabular}{@{}ccc@{}}
        \includegraphics[width=0.25\textwidth]{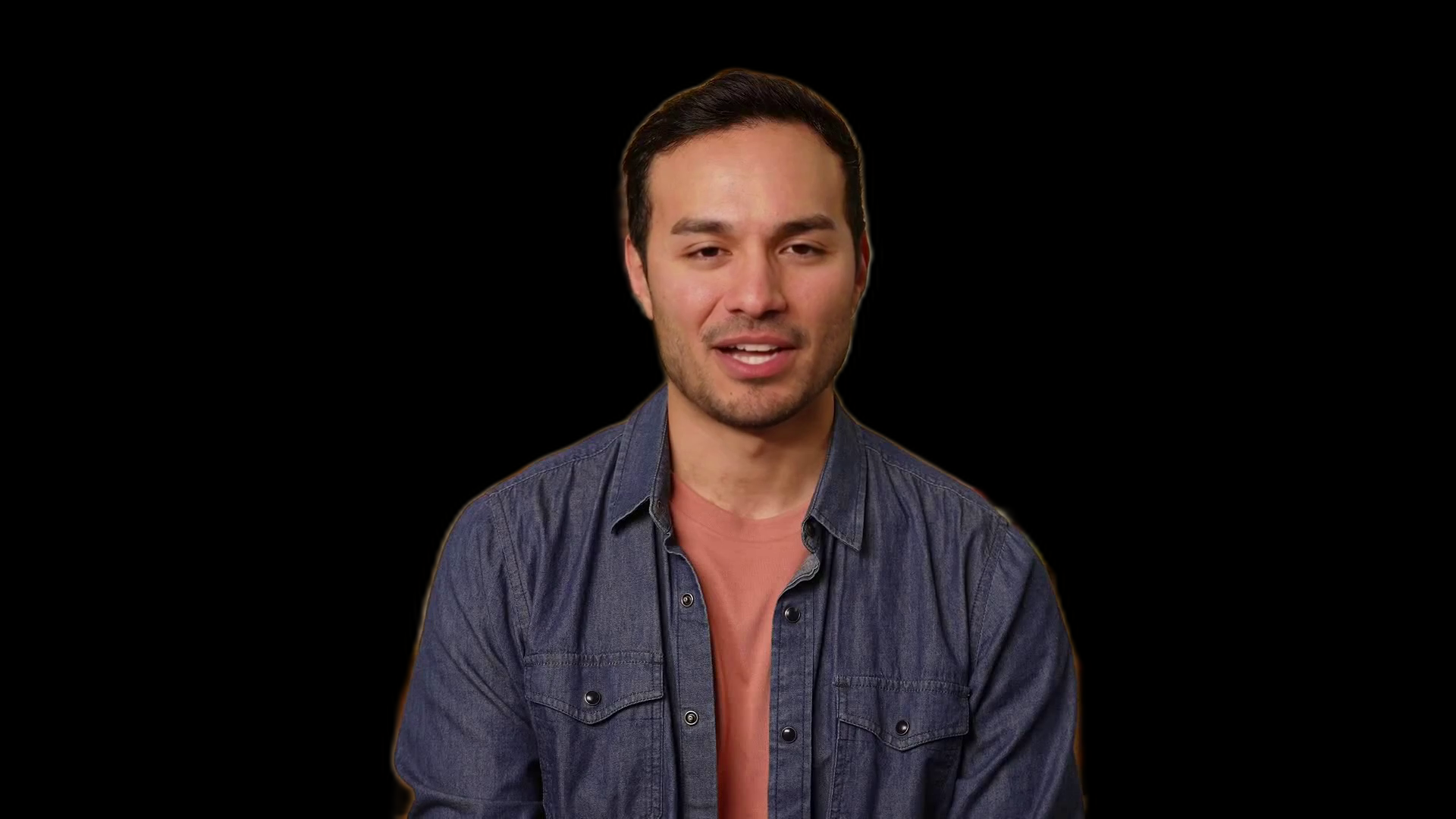} &
        \includegraphics[width=0.25\textwidth]{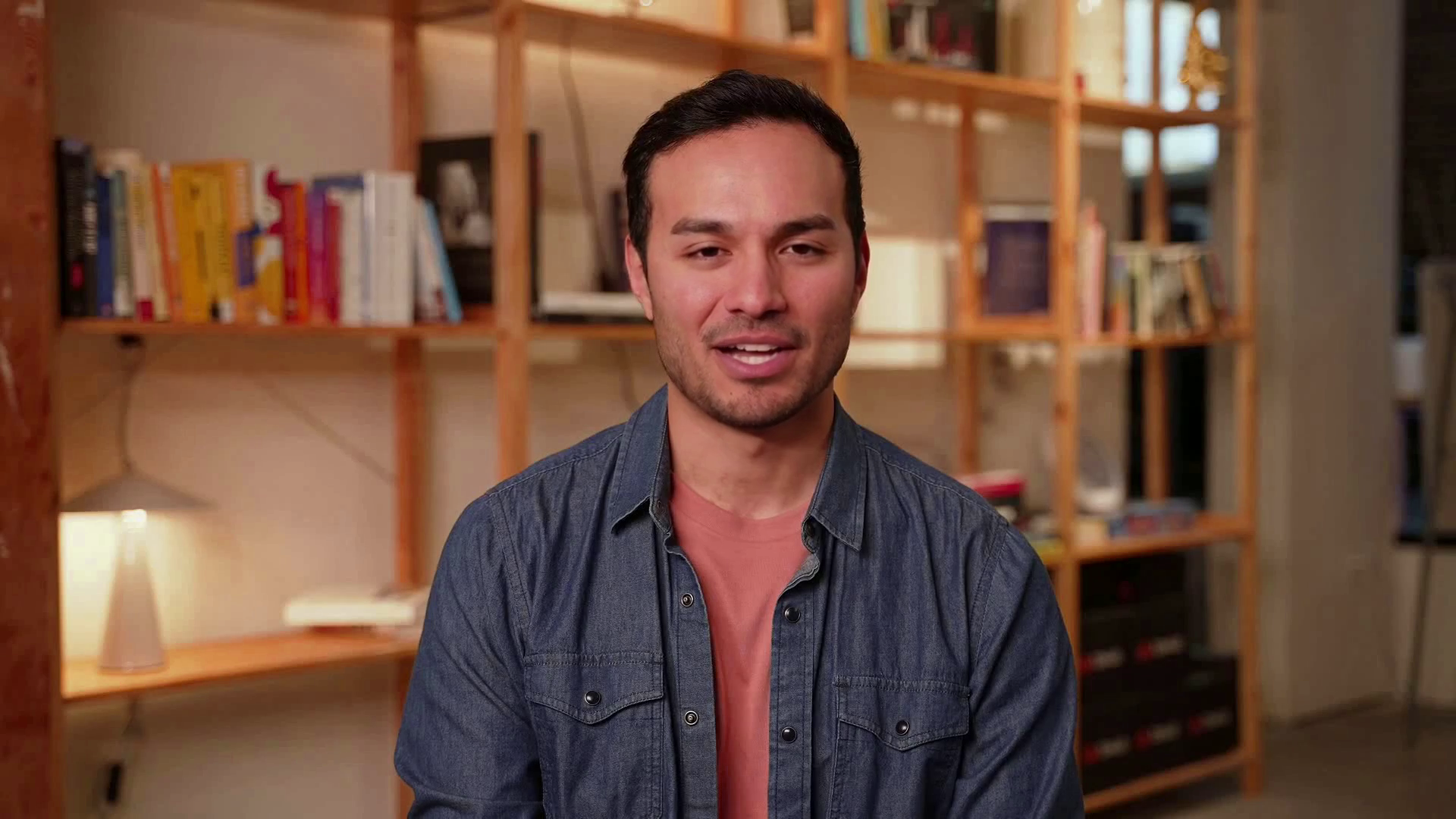} &
        \includegraphics[width=0.25\textwidth]{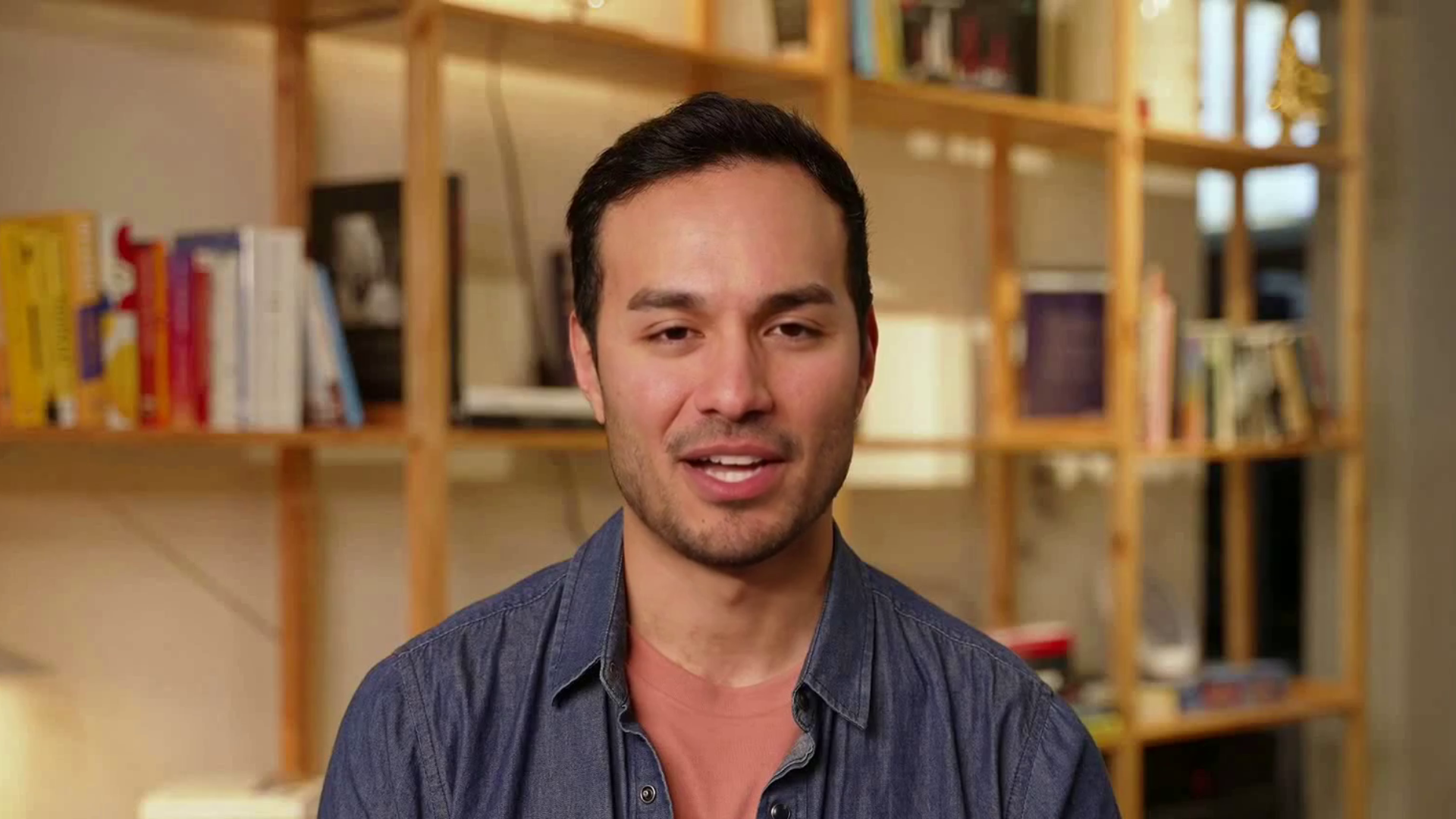} \\
        {\small (a) Background removal} & {\small (b) Colour} & {\small (c) Zoom} \\
    \end{tabular}
    \vspace{-0.2cm}
    \caption{\textit{Example attacks from our benchmark. Background removal and zoom are common in avatar workflows but challenging for existing watermarking methods.}}
    \label{fig:attacks}
\end{figure*}

\begin{figure*}[t]
    \centering
    \setlength{\tabcolsep}{4pt}
    \begin{tabular}{@{}cc@{}}
        \includegraphics[width=0.35\textwidth]{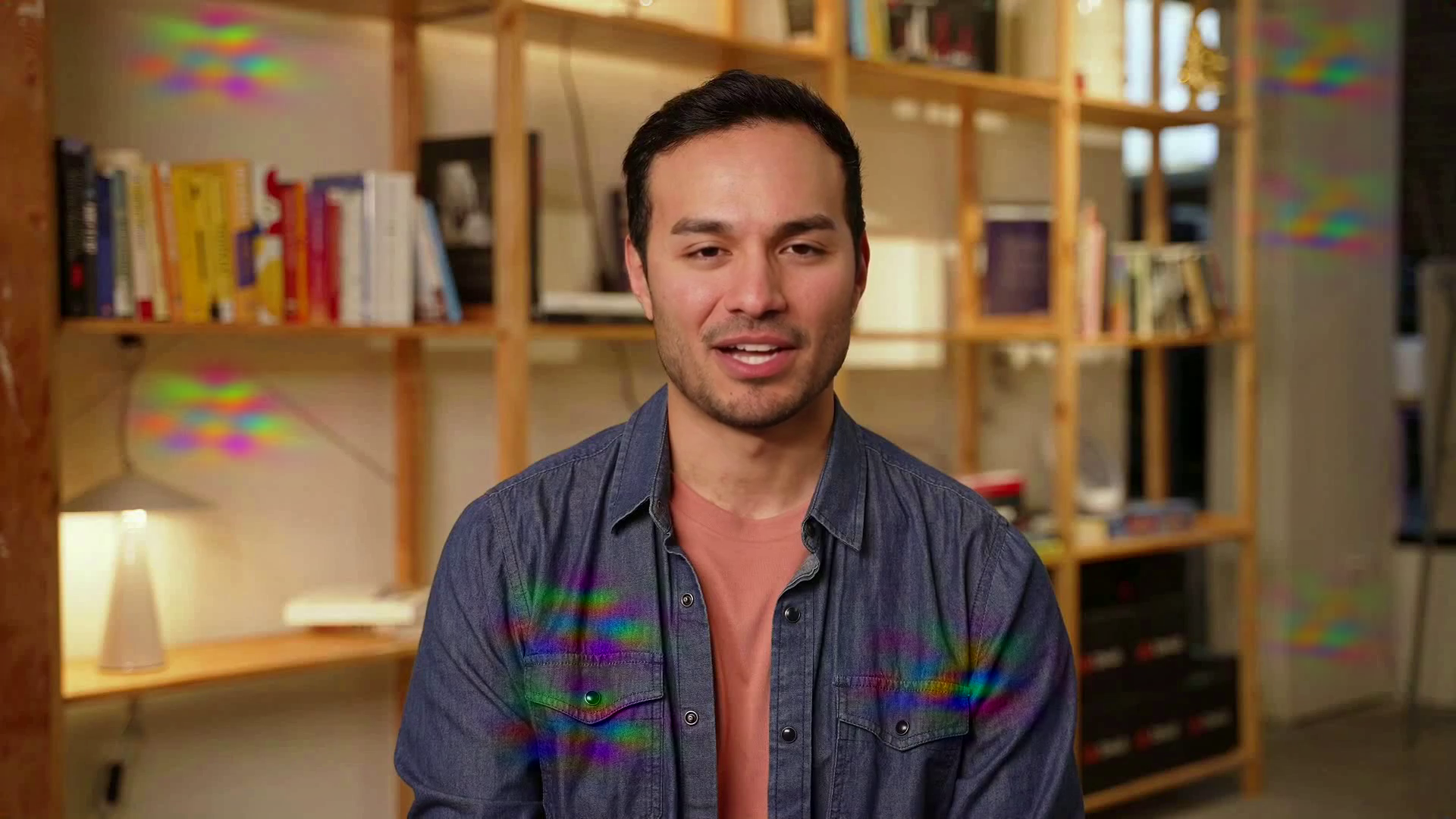} &
        \includegraphics[width=0.35\textwidth]{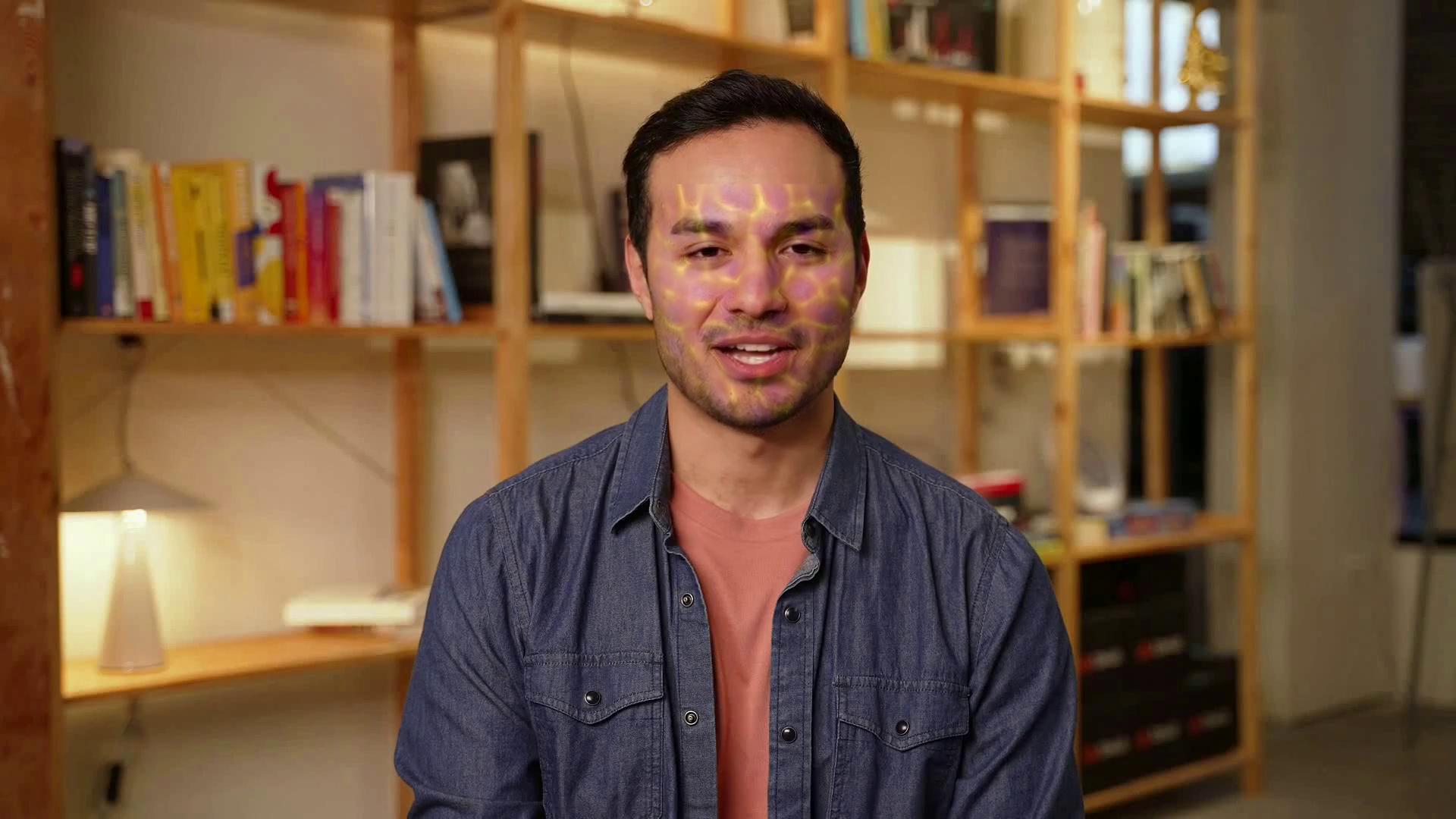} \\
        {\small (a) VideoSeal} & {\small (b) WALT (ours)} \\
    \end{tabular}
    \vspace{-0.2cm}
    \caption{\textit{Watermark visualisation (amplified $\approx10\times$ for visibility). VideoSeal embeds across the entire frame; WALT embeds only on the face via UV texture mapping, providing inherent robustness to background removal and cropping.}}
    \label{fig:watermark}
\end{figure*}
\section{Related Work}
\label{sec:related}

\textbf{Image Watermarking.}
Deep learning approaches to image watermarking encode binary messages into images via encoder-decoder networks trained end-to-end. HiDDeN~\cite{hidden} pioneered this approach with adversarial training for imperceptibility. Subsequent work improved robustness: MBRS~\cite{mbrs} uses mini-batch training with diverse augmentations, CIN~\cite{cin} employs invertible networks, TrustMark~\cite{trustmark} optimises for standards compliance, and WAM~\cite{wam} focuses on social media compression. RoSteALS~\cite{rosteals} achieves state-of-the-art quality by embedding in pretrained autoencoder latent space. However, these methods operate frame-by-frame without temporal consistency.

\textbf{Video Watermarking.}
VideoSeal~\cite{videoseal} extends image watermarking to video with temporal augmentations and 3D convolutions for improved robustness. While effective for general video, it does not account for avatar-specific transformations such as background replacement or face-centred reframing.

\textbf{3D Face Reconstruction.}
Morphable face models enable dense correspondence between face images and canonical UV space. FLAME~\cite{flame} provides a differentiable face model, DECA~\cite{deca} enables single-image reconstruction, and EMOCA~\cite{emoca} improves expression capture. We leverage these advances to embed watermarks in UV texture space, ensuring geometric consistency.

\section{RAW Benchmark}
\label{sec:benchmark}

To accurately benchmark the ability of watermarking methods to work on generated avatars, we propose a novel dataset. This dataset consists of 50 videos from 5 commercial companies specialising in avatar generation: D-ID~\cite{d-id}, HeyGen~\cite{heygen}, Synthesia~\cite{synthesia}, Tavus~\cite{tavus}, and Vidnoz~\cite{vidnoz}. These companies employ many different state-of-the-art methods to generate avatars from text input. For each company, we generate 10 videos using phonetic pangrams to ensure diversity of face movements. While 50 videos is modest in scale, it spans 5 providers with diverse generation methods and is sufficient to reveal significant performance gaps between watermarking approaches.

We test robustness by applying 6 attacks to watermarked videos, simulating real-world avatar post-processing (see Figure~\ref{fig:attacks}):
\begin{itemize}
\item \textbf{Background removal} via \texttt{rembg}~\cite{rembg}
\item \textbf{Colour} transformations in HSV space (brightness/contrast $\times 0.8$-$1.2$, hue $\pm 10^{\circ}$, saturation $\times 0.8$-$1.2$)
\item \textbf{Crop} to a face-centred region via MediaPipe~\cite{mediapipe} (ratio $0.6$-$0.9$), reducing output resolution;
\item \textbf{Framerate} subsampling to 10-50\,fps;
\item \textbf{MP4 compression} via H.264 re-encoding (CRF 20--30);
\item \textbf{Zoom} with face-centred $1.2$-$1.8\times$ magnification, cropping around the face and resizing back to the original resolution.
\end{itemize}
Attack parameters are sampled randomly within the specified ranges using a fixed seed, ensuring identical transformations are applied across all methods for fair comparison.

\begin{table*}[t]
\centering
\caption{\textit{Robustness to attacks (bit accuracy). Best in} \textbf{\textit{bold}}\textit{, second-best} \underline{\textit{underlined}}\textit{.}}
\label{tab:attacks}
\resizebox{\textwidth}{!}{%
\begin{tabular}{llcccccc}
\toprule
& Method & Background $\uparrow$ & Colour $\uparrow$ & Crop $\uparrow$ & Framerate $\uparrow$ & MP4 $\uparrow$ & Zoom $\uparrow$ \\
\midrule
\multirow{6}{*}{\rotatebox{90}{Image}}
& Hidden~\cite{hidden} & 0.888 \scriptsize{±.036} & 0.957 \scriptsize{±.040} & 0.967 \scriptsize{±.035} & 0.985 \scriptsize{±.019} & 0.985 \scriptsize{±.017} & 0.639 \scriptsize{±.083} \\
& MBRS~\cite{mbrs} & 0.606 \scriptsize{±.042} & 0.941 \scriptsize{±.063} & 0.749 \scriptsize{±.217} & 0.965 \scriptsize{±.042} & 0.962 \scriptsize{±.039} & 0.504 \scriptsize{±.033} \\
& CIN~\cite{cin} & \textbf{0.965} \scriptsize{±.051} & \textbf{1.000} \scriptsize{±.000} & 0.729 \scriptsize{±.289} & \textbf{1.000} \scriptsize{±.000} & \textbf{1.000} \scriptsize{±.000} & 0.509 \scriptsize{±.083} \\
& WAM~\cite{wam} & 0.916 \scriptsize{±.060} & \underline{0.998} \scriptsize{±.009} & \underline{0.959} \scriptsize{±.056} & \underline{0.998} \scriptsize{±.010} & \underline{0.998} \scriptsize{±.007} & \underline{0.888} \scriptsize{±.085} \\
& TrustMark~\cite{trustmark} & 0.750 \scriptsize{±.064} & 0.960 \scriptsize{±.077} & \textbf{0.997} \scriptsize{±.008} & 0.997 \scriptsize{±.009} & \underline{0.998} \scriptsize{±.007} & 0.524 \scriptsize{±.103} \\
& RoSteALS~\cite{rosteals} & 0.734 \scriptsize{±.045} & 0.996 \scriptsize{±.006} & 0.783 \scriptsize{±.227} & \underline{0.998} \scriptsize{±.004} & \underline{0.998} \scriptsize{±.004} & 0.498 \scriptsize{±.034} \\
\midrule
\multirow{1}{*}{\rotatebox{90}{\scriptsize{Video}}}
& VideoSeal~\cite{videoseal} & 0.799 \scriptsize{±.173} & 0.947 \scriptsize{±.050} & 0.951 \scriptsize{±.043} & 0.966 \scriptsize{±.032} & 0.965 \scriptsize{±.032} & 0.741 \scriptsize{±.134} \\
\midrule
\multirow{1}{*}{\rotatebox{90}{\scriptsize{Face}}}
& WALT (ours) & \underline{0.956} \scriptsize{±.026} & 0.948 \scriptsize{±.028} & 0.960 \scriptsize{±.025} & 0.956 \scriptsize{±.049} & 0.870 \scriptsize{±.034} & \textbf{0.924} \scriptsize{±.030} \\
\bottomrule
\end{tabular}%
}
\end{table*}

\section{WALT: Watermarking Avatars with Learned Textures}
\label{sec:method}

\begin{figure}[t]
    \centering
    \includegraphics[width=0.5 \textwidth]{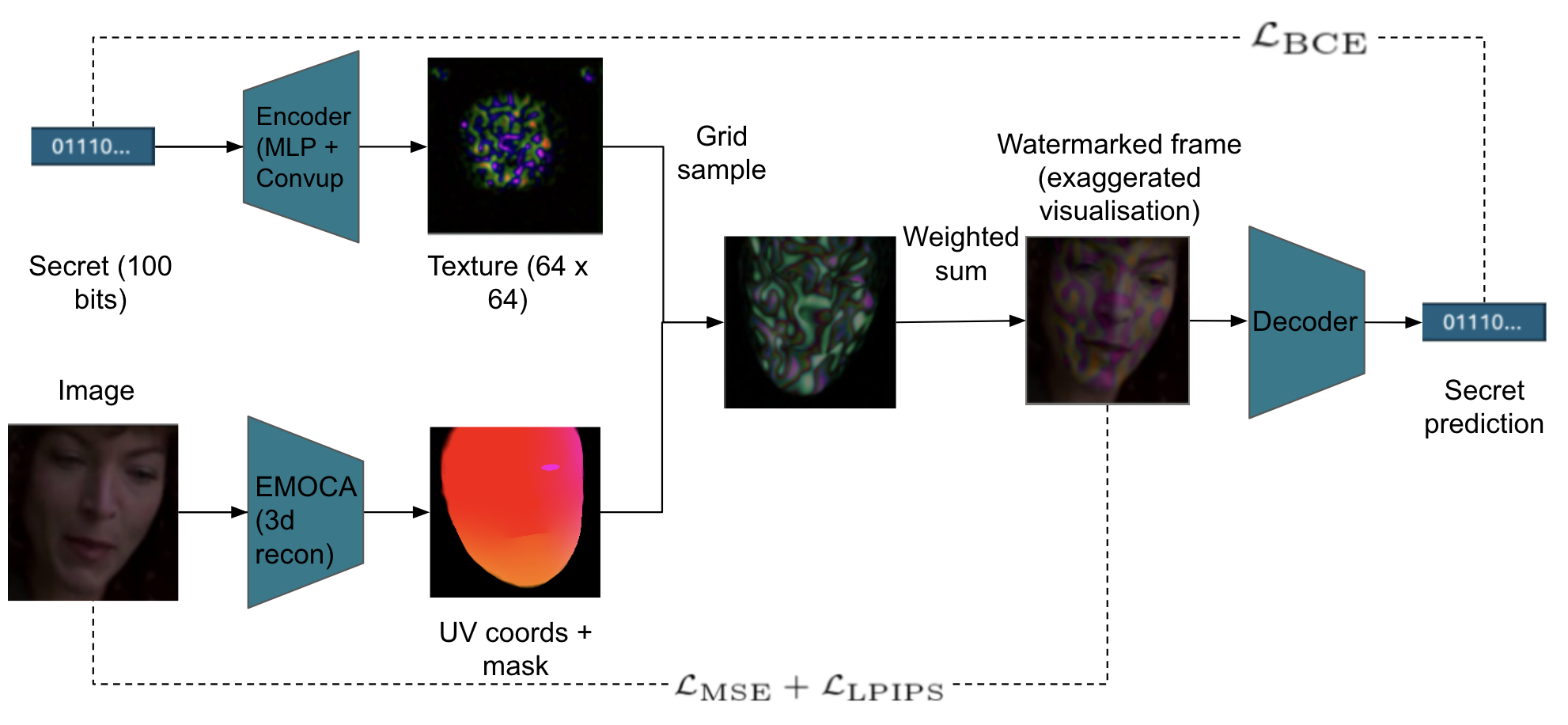}
    \caption{\textit{WALT architecture. The encoder transforms a 100-bit secret into a learnable texture. EMOCA reconstructs 3D face geometry to obtain UV coordinates, which are used to sample the texture onto the face region via grid sampling. The watermarked frame is produced by weighted addition. A ResNet-50 decoder extracts the secret from the watermarked image. Training uses BCE loss for bit accuracy and MSE + LPIPS losses for visual quality.}}
    \label{fig:architecture}
\end{figure}

We propose WALT, a watermarking method designed specifically for digital avatars. Unlike existing approaches that embed watermarks directly into pixel space, WALT operates in UV texture space, ensuring the watermark follows facial geometry and remains temporally consistent across video frames.

Our pipeline consists of three components: (1) a \textbf{secret encoder} that transforms a binary message into a learnable texture, (2) a \textbf{texture sampler} that applies this texture to the face region using 3D face reconstruction, and (3) a \textbf{decoder} that extracts the message from watermarked frames.

Given a binary secret $s \in \{0,1\}^{L}$ of length $L=100$ bits, the encoder $E$ generates a watermark texture $T \in \mathbb{R}^{3 \times 256 \times 256}$. The encoder first projects $s$ to a $64 \times 64 \times 64$ feature map via a linear layer with SiLU activation, concatenates 2D positional encodings (2 channels), then applies two upsampling blocks (each comprising bilinear $2\times$ upsampling followed by two $3\times3$ convolutional layers with SiLU activations), reducing channels $66 \to 32 \to 16$. A final $1\times1$ convolution with tanh activation produces the 3-channel texture.

To apply the watermark, we leverage 3D Morphable Face Models. Given an input frame $I$, we use EMOCA~\cite{emoca}, which builds on DECA~\cite{deca} and the FLAME~\cite{flame} face model, to reconstruct facial geometry and obtain dense UV coordinates $U \in \mathbb{R}^{H \times W \times 2}$ mapping each face pixel to texture space, along with a visibility mask $M$.

The watermark is applied by sampling the texture at the UV coordinates and adding it to the original image:

\begin{equation}
    I_w = I + w \cdot (\text{sample}(T, U) \odot M)
\end{equation}

where $w$ controls watermark strength (set to $0.03$) and $\odot$ denotes element-wise multiplication. This formulation ensures the watermark is applied \textit{only} to the face region, providing inherent robustness to background modifications. The UV mapping is determined by facial geometry rather than pixel position, ensuring the same facial pose produces consistent UV coordinates across frames. This produces a watermark that moves naturally with the face, eliminating temporal flickering common in frame-by-frame approaches. Figure~\ref{fig:watermark} illustrates the key difference: VideoSeal embeds watermarks across the entire frame, including background and clothing, while WALT restricts embedding to the face region. This explains WALT's robustness to background removal, where the watermark remains intact even when the background is stripped.

For decoding, we use a ResNet-50~\cite{resnet} that takes the full watermarked image (resized to $256 \times 256$) and directly predicts the $L=100$ embedded bits via $\hat{s} = D(I_w)$.
The decoder operates on the raw image without requiring face reconstruction at decode time. This improves robustness even under geometric distortions where precise UV recovery would fail.

We adapt training from RoSteALS~\cite{rosteals}, replacing their VQGAN with our UV texture space while retaining their loss formulation and noise augmentation schedule. The encoder and decoder are trained end-to-end on the MIRFlickR~\cite{mirflickr} image dataset using reconstruction losses (MSE + LPIPS~\cite{lpips}) for visual quality and binary cross-entropy for bit accuracy. The RAW benchmark videos are used exclusively for evaluation, meaning WALT generalises zero-shot to avatars unseen during training.

\vspace{-0.2cm}
\section{Experiments}
\label{sec:experiments}

\begin{table}[t]
\centering
\caption{\textit{Visual quality metrics. Methods vary in message capacity (30--256 bits); higher capacity increases encoding/decoding difficulty. Best results in} \textbf{\textit{bold}}\textit{.}}
\label{tab:quality}
\resizebox{\columnwidth}{!}{%
\begin{tabular}{llrcccc}
\toprule
& Method & Bits & Accuracy $\uparrow$ & PSNR $\uparrow$ & SSIM $\uparrow$ & LPIPS $\downarrow$ \\
\midrule
\multirow{6}{*}{\rotatebox{90}{Image}}
& Hidden~\cite{hidden} & 48 & 0.992 \scriptsize{±.014} & 32.38 \scriptsize{±1.10} & 0.873 \scriptsize{±.055} & 0.093 \scriptsize{±.052} \\
& MBRS~\cite{mbrs} & 256 & 0.982 \scriptsize{±.020} & 45.31 \scriptsize{±1.01} & 0.991 \scriptsize{±.003} & 0.005 \scriptsize{±.004} \\
& CIN~\cite{cin} & 30 & \textbf{1.000} \scriptsize{±.000} & 44.59 \scriptsize{±0.86} & 0.989 \scriptsize{±.006} & 0.005 \scriptsize{±.004} \\
& WAM~\cite{wam} & 32 & 0.998 \scriptsize{±.010} & 42.53 \scriptsize{±1.60} & 0.981 \scriptsize{±.020} & 0.009 \scriptsize{±.006} \\
& TrustMark~\cite{trustmark} & 100 & 0.999 \scriptsize{±.006} & 43.07 \scriptsize{±1.60} & 0.988 \scriptsize{±.009} & \textbf{0.002} \scriptsize{±.001} \\
& RoSteALS~\cite{rosteals} & 100 & 0.998 \scriptsize{±.004} & 31.26 \scriptsize{±4.31} & 0.922 \scriptsize{±.070} & 0.025 \scriptsize{±.020} \\
\midrule
\multirow{1}{*}{\rotatebox{90}{\scriptsize{Video}}}
& VideoSeal~\cite{videoseal} & 96 & 0.973 \scriptsize{±.030} & \textbf{46.61} \scriptsize{±1.16} & {0.993} \scriptsize{±.006} & 0.004 \scriptsize{±.003} \\
\midrule
\multirow{1}{*}{\rotatebox{90}{\scriptsize{Face}}}
& WALT (ours) & 100 & 0.960 \scriptsize{±.024} & 45.71 \scriptsize{±3.79} & \textbf{0.995} \scriptsize{±.012} & \textbf{0.002} \scriptsize{±.006} \\
\bottomrule
\end{tabular}%
}
\end{table}

We evaluate using four metrics: \textbf{Accuracy} measures the fraction of correctly recovered bits across all $L=100$ message bits, averaged over frames and then over videos. Visual quality is assessed via \textbf{PSNR} (peak signal-to-noise ratio), \textbf{SSIM} (structural similarity), and \textbf{LPIPS}~\cite{lpips} (learned perceptual similarity, lower is better).

Table~\ref{tab:quality} presents visual quality metrics. All approaches achieve high perceptual quality, with WALT matching the best SSIM and LPIPS scores while encoding 100 bits. Note that CIN achieves perfect accuracy but encodes only 30 bits.

Table~\ref{tab:attacks} shows robustness under our attack suite. Zoom proves the most challenging attack, with most methods falling below 75\% accuracy; WALT achieves the best zoom robustness (92.4\%) by operating in facial geometry space. WALT also achieves the second-best background removal robustness (95.6\%), behind only CIN.

We also experimented with a simpler baseline: applying VideoSeal only to a cropped face region. While this achieved strong background (97.5\%) and zoom (97.7\%) robustness, it failed catastrophically under MP4 compression (69.1\%), likely because concentrating watermark information in a smaller spatial region makes it disproportionately vulnerable to compression artifacts. WALT's learned texture representation avoids this failure mode, maintaining 87.0\% compression robustness.

\section{Discussion and Conclusion}
\label{sec:conclusion}

We introduced the RAW benchmark for evaluating watermarking on digital avatars. Our evaluation reveals that avatar-specific attacks such as background removal and zoom pose significant challenges for existing methods. Face-localised watermarking addresses these challenges, but a naive face-cropping baseline (applying VideoSeal to a cropped face region) failed catastrophically on compression (69.1\%), likely because concentrating watermark information in a smaller spatial region makes it disproportionately vulnerable to compression artifacts. WALT's learned texture approach recovers compression robustness (87.0\%) by distributing watermark information more robustly across the face in UV space.

WALT's clean accuracy (96.0\%) is lower than some baselines, and compression robustness remains below methods like CIN, which,while only embedding 30 bits instead of WALT's 100, leverages an invertible network architecture which likely distributes information across the full frequency spectrum. Future work includes quantitative temporal consistency evaluation, end-to-end video training, additional attacks such as lip-sync modification, comparison with recent methods such as VINE~\cite{vine}, and analysis of failure cases from face tracking errors (e.g.\ occlusions or extreme poses). We release our dataset at \texttt{zenodo.org/records/20247264} (doi:10.2312/egs.20261006) and code and model weights at \texttt{github.com/Jack-Paz/RAW-Benchmark}.

\vspace{-0.2cm}

\bibliographystyle{eg-alpha-doi}
\bibliography{egbibsample}

\end{document}